\documentclass[11pt,a4paper]{article}
\usepackage[hyperref]{acl2021}

\usepackage{times}
\usepackage{latexsym}

\usepackage{times}
\usepackage{latexsym}
\usepackage{amsmath}
\usepackage{url}
\usepackage{mathdots}
\usepackage{yhmath}
\usepackage{cancel}
\usepackage{color}
\usepackage{siunitx}
\usepackage{array}
\usepackage{multirow}
\usepackage{amssymb}
\usepackage{gensymb}
\usepackage{tabularx}
\usepackage{booktabs}
\usepackage{svg}
\usepackage{tablefootnote}
\usepackage{cleveref}
\usepackage{tabularx}
\usepackage{xcolor}
\usepackage{makecell}
\usepackage{enumitem}
\usepackage{amsfonts}
\usepackage{pifont}
\usepackage{arydshln}

\usepackage[T1]{fontenc}

\usepackage[utf8]{inputenc}

\aclfinalcopy

\usepackage{microtype}

\newcommand{\ours}{\textsc{Scarce}}

\newcommand{\oursSingle}{\textsc{Scarce-Single}}
\newcommand{\oursMulti}{\textsc{Scarce-Multi}}

\newcommand{\commonsense}{\textsc{CommonSense}}
\newcommand{\paraphrase}{\textsc{Paraphrase}}
\newcommand{\single}{\textsc{Single}}
\newcommand{\multiref}{\textsc{Multi}}
\newcommand{\comet}{COMET}

\title{Improving Automated Evaluation of Open Domain Dialog \\via Diverse Reference Augmentation} 

\author{Varun Gangal \thanks{\quad VG and HJ contributed equally for this paper. Order decided by coin flip.} $^1$ \quad Harsh Jhamtani * $^1$ \quad Eduard Hovy $^1$ \quad Taylor Berg-Kirkpatrick $^2$ \\
$^1$ School of Computer Science, Carnegie Mellon University\\
$^2$ Computer Science and Engineering. University of California San Diego \\
\tt{\{vgangal,jharsh,hovy\}@cs.cmu.edu, tberg@ucsd.eng.edu}
}

\begin{document}
\maketitle

\begin{abstract}

Multiple different responses are often plausible for a given open domain dialog context.
Prior work has shown the importance of having multiple valid reference responses for meaningful 
and robust
automated evaluations. In such cases, common practice has been to collect more human written references. 
However, such collection can be expensive, time consuming, and not easily scalable. 
Instead, we propose a novel technique for \emph{automatically} expanding a human generated reference to a set of candidate references. We fetch plausible references from knowledge sources, and adapt them so that they are more fluent in context of the dialog instance in question. More specifically, we use (1) a commonsense knowledge base to elicit a large number of plausible reactions given the dialog history (2) relevant instances retrieved from dialog corpus, using similar past as well as future contexts. 
We demonstrate that our automatically expanded reference sets lead to large improvements in correlations of automated metrics with human ratings of system outputs for DailyDialog dataset. \footnote{Code and data are available at \url{https://github.com/harsh19/Diverse-Reference-Augmentation/} }

\end{abstract}


\section{Introduction}

Evaluation by human annotators perhaps give the best insights into quality of machine generated natural language outputs. However, they can be expensive and time consuming. Much focus has therefore been on automated evaluation methods which correlate with human evaluations. Automated metrics such as BLEU \cite{papineni2002bleu} 
work well for tasks such as machine translation, but often correlate poorly with human ratings in tasks such as open domain dialog which admit a wide variety of valid response for given context, often due to small number of human written references \cite{ZhaoZE17,sai2020survey}. Prior work \cite{sugiyama2019automatic,gupta2019investigating} has demonstrated that having multiple valid references for the same context leads to automated metrics being better correlated to human judgements for appropriateness.  However, collecting human written responses is difficult to scale, can be costly, and may find it difficult to 
cover a large variety of correct responses \cite{celikyilmaz2020evaluation}.

\begin{figure}[t]
    \centering
    \includegraphics[width=0.48\textwidth]{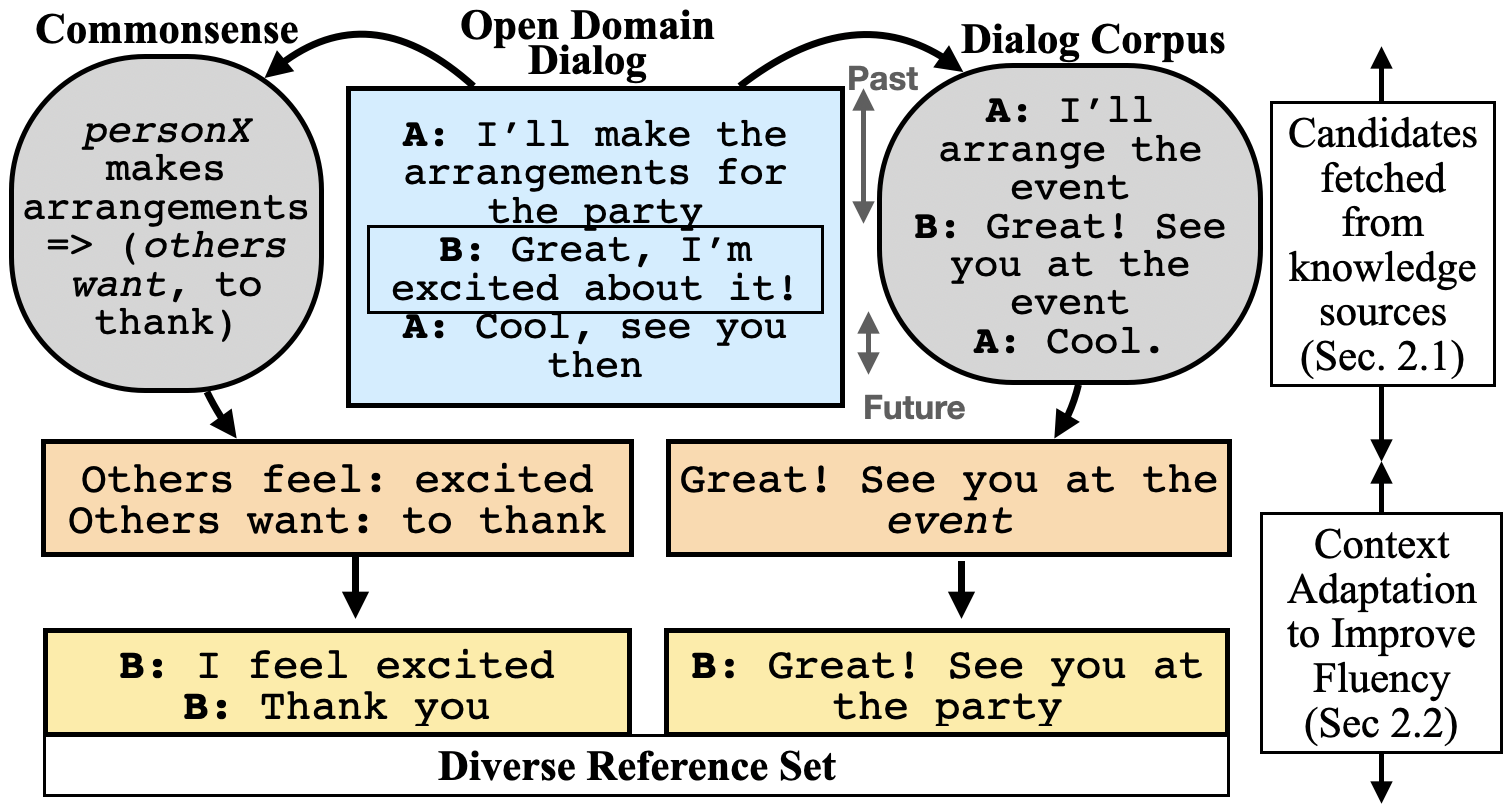} 
    \caption{\small We propose automatic ways to collect references sans any crowd-sourcing, through two types of knowledge sources: commonsense 
    and retrieved instance 
    knowledge, followed by automated adaptation to make them more fluent in the target contexts.}
    \label{fig:overview}
    \vspace{-3ex}
\end{figure}

In this work, we automatically extract a large number of diverse references to be used with such reference-based metrics, without resorting to expensive crowd-sourcing. Intuitively, since open-domain dialog pertains to everyday life, its utterance text tends to re-instantiate from a large but limited pool of situations \cite{schank1972conceptual} e.g friends debating politics etc, with variation only on some details e.g country discussed. Hence, knowledge encapsulating a wide scope of situations can serve as one starting point to automatically seed a set of diverse references. 
We first fetch plausible candidates from two types of 
knowledge sources (Figure \ref{fig:overview}). 
Such knowledge sources provide ready and easy access to a large number of potentially appropriate and diverse references.
However, all retrieved instances may not be directly useful. 
As such, to achieve more fluent references, we propose techniques to adapt the candidate references based on the context (e.g change country being discussed).
Note that since we are interested in creating references for only evaluating appropriateness of system outputs, our techniques can rely on broader data sources compared to dialog models. For example, we use future context 
and human written reference
for retrieval, while a dialog model cannot.

Our contributions are as follows:
(1) We propose a method for automated reference set augmentation for automated dialog evaluation. Compared to collecting more human-written responses, our approach is inexpensive and scalable, and fetches a diverse set of references. 
(2) We observe high correlations of various automated metrics with human ratings when proposed reference augmentation is applied to the test split of DailyDialog dataset \cite{li-etal-2017-dailydialog}.
We additionally observe that paraphrasing, a popular data augmentation technique, performs much worse.
(3) We employ novel use of commonsense knowledge and dialog corpus instances, and unsupervised techniques for adapting retrieved references into more fluent forms.

\section{Method}

Figure \ref{fig:overview} shows an overview of our proposed methodology. We first fetch plausible candidates from two types of 
knowledge sources. 
Thereafter, the retrieved candidate references are adapted so that they are fluent in the target context.
We refer to our proposed method as \textbf{\ours{}} ( 
\textbf{SC}alable \textbf{A}utomated \textbf{R}eference \textbf{C}onstruction for \textbf{E}valuation).

\subsection{Knowledge Sources}
\noindent \textbf{Pre-trained Commonsense Model}
\label{subsec:commonsense}
Much open domain dialog is based on everyday matters.
We posit that extracting inferences about a situation using a commonsense knowledge base could be useful in identifying a wide variety of plausible reactions for a given dialog context. For example, a person making arrangements for an event might receive thanks from others (Figure \ref{fig:overview}).
We utilize \comet{} \cite{BosselutRSMCC19} 
an off-the-shelf commonsense knowledge model built on ATOMIC \cite{SapBABLRRSC19} or ConceptNet \cite{speer2017conceptnet} corpus. 
It can be used to elicit
commonsense inferences.

\comet{}-ATOMIC provides inferences on cause-effect interrelations between events pertaining to nine relation types such as \emph{oReact} (effect on others due to the event), and \emph{oWant} (inferences about wants of the receiver of event). 
Given an utterance from the previous speaker, we draw up to 5 inferences pertaining to each of oEffect, oReact, and oWant relation types to construct plausible references for the target response. For example, for an utterance `I will make the arrangements. It will be great.', one of the inferences corresponding to oEffect is `feel excited', depicting a plausible state of the next dialog speaker. 
However, such outputs are typically phrases, and we discuss transformation to fluent sentences in Section \ref{sec:context}. 
Similarly, we use inferences pertaining to `CausesDesire' and `HasFirstSubevent' relation types  from \comet{}-ConceptNet. 
\\

\noindent \textbf{Dialog Corpus Retrieval}
For a test dialog context under consideration, one is likely to find similar contexts occurring in some of the training dialogues, given a sufficient number of them. Using retrieval, we can identify such contexts and use their responses as pseudo-references for the test-time response. 
Specifically, for retrieval, we use the BM25 function $S_{bm25}(x,y)$ \cite{robertson1995okapi}  to compare each element $\{d^{past}_{t},d^{resp}_{t},d^{future}_{t}\}$ of the turn under evaluation $d_{t}$ (the query) with those of the candidate turn $x_{t'}$, $\{x^{past}_{t'},x^{resp}_{t'},x^{future}_{t'}\}$. Here, $d^{past}_{t}$ and $d^{future}_t$ are the windows of turn sequences before and after response $d^{resp}_{t}$. 

Our approach is related to \citet{galley2015deltableu}, who propose $\Delta$-BLEU measure which uses retrieval to produce pseudo-references. However, unlike here, they require annotator quality scores to weigh them during evaluation. Moreover, though we utilize retrieval for evaluation, methods of this kind have found success in many generation setups \cite{li2018delete,peng2019text,khandelwal2019generalization}.
Besides being automatic, our method differs from the above ones in that it explores the added utility of future information for retrieval. For instance, for the dialog shown in Figure 1, besides matching ``Great!'' in the response, our retrieval benefits from matching ``cool'' in the future.

\subsection{Context Adaptation}
\label{sec:context}
We note that commonsense knowledge outputs are incomplete sentences, and we use simple templates to convert them to fluent sentences e.g. `feels excited' gets transformed to `i feel excited'. (Detailed templates in Appendix B)
Further, we note that references from knowledge sources are often not fluent for the target context. For example, `event' in the retrieved reference shown in Figure \ref{fig:overview} can be updated to `party' to construct a more apt reference. To adapt the retrieved text to better fit the target context we use employ an unsupervised decoding procedure, based on the approach of \citet{DBLP:conf/emnlp/QinSWBHBBC20}, that uses gradient ascent to search for output text that maximizes 
 (1) fluency with the left context (approximated by the likelihood of the output text under a pretrained GPT-2 model) and (2) similarity to the original text from the knowledge source (approximated by the likelihood of the original text under the output text's token-level word distributions). The method utilizes a heuristic update procedure to iteratively refine a differentiable proxy for the output text (a sequence token-level word distributions), while keeping the model parameters fixed. More details can be found in \citet{DBLP:conf/emnlp/QinSWBHBBC20} and in Appendix B.

\begin{table*}[t]
\centering
\small 
\begin{tabular}{@{}l||c|c|c||c|c|c@{}}
  \multicolumn{7}{c}{Spearman Rank Correlation / Kendall Tau Rank Correlation} \\ \toprule
  Setup  &  \multicolumn{3}{c||}{1 human written reference} &  \multicolumn{3}{c}{4 human written references}  \\ \midrule
    Dataset  & \multicolumn{1}{c|}{\single{}} & \paraphrase{} & \multicolumn{1}{c||}{\ours{}}  & \multicolumn{1}{c|}{\multiref{}} & \paraphrase{} & \multicolumn{1}{c}{\ours{}} \\ 
   & \multicolumn{1}{c|}{\tiny{\cite{li-etal-2017-dailydialog}}} & -\single{} & \multicolumn{1}{c||}{-\single{}(Ours)} & \multicolumn{1}{c|}{\tiny{\cite{gupta2019investigating}}} & -\multiref{} & \multicolumn{1}{c}{-\multiref{}(Ours)} \\ \midrule
BLEU-4 & 0.09 / 0.07 & 0.13 / 0.09 & 0.30 / 0.21 & 0.28 / 0.20 & 0.27 / 0.19 & \textbf{0.36} / 0.25 \\
BLEU-3 & 0.06 / 0.04 & 0.11 / 0.07 & 0.29 / 0.20 & 0.24 / 0.17 & 0.24 / 0.17 & 0.35 / 0.24 \\
BLEU-2 & 0.04 / 0.03 & 0.08 / 0.06 & 0.28 / 0.19 & 0.20 / 0.14 & 0.21 / 0.14 & 0.33 / 0.23 \\
BLEU-1 & 0.02 / 0.02 & 0.06 / 0.04 & 0.25 / 0.17 & 0.19 / 0.13 & 0.18 / 0.12 & 0.29 / 0.21 \\
ROUGE-L & 0.07 / 0.05 & 0.09 / 0.06 & 0.26 / 0.18 & 0.20 / 0.14 & 0.20 / 0.14 & 0.32 / 0.22 \\
METEOR & 0.11 / 0.07 & 0.09 / 0.06 & 0.24 / 0.17 & 0.23 / 0.16 & 0.22 / 0.15 & 0.30 / 0.21 \\
\midrule
EmbeddingAvg & 0.03 / 0.02 & 0.02 / 0.01 & 0.02 / 0.02 & 0.10 / 0.07 & 0.10 / 0.07 & 0.08 / 0.05 \\
SkipThought & -0.00 / 0.00 & -0.03 / -0.02 & 0.09 / 0.07 & 0.07 / 0.05 & 0.05 / 0.04 & 0.13 / 0.10 \\
BERT-Prec & 0.27 / 0.19 & 0.28 / 0.19 & 0.38 / 0.26 & 0.32 / 0.22 & 0.32 / 0.22 & \textbf{0.41} / 0.28 \\
BERT-Rec & 0.10 / 0.06 & 0.09 / 0.06 & 0.24 / 0.16 & 0.23 / 0.16 & 0.21 / 0.15 & 0.30 / 0.21 \\
\midrule
Max. value & 0.27 / 0.19 & 0.28 / 0.19 & 0.38 / 0.26 & 0.32 / 0.22 & 0.32 / 0.22 & 0.41 / 0.28 \\
\bottomrule
\end{tabular}
\vspace{-0.2\abovedisplayskip}
\caption{
Utterance level Spearman Rank Correlation \shortcite{spearman1961proof} and Kendall Tau Rank Correlations \shortcite{kendall1938new}. 
(1) \oursSingle{} augments the original single human written response (\single{}) in DailyDialog dataset \cite{li-etal-2017-dailydialog}  using proposed method. It leads to large improvements in correlations across most of the metrics, when compared to \single{}. 
(2) \oursMulti{} augments the \multiref{} dataset, again leading to improvements in correlations to human ratings, especially for BLEU and BERT-Prec metrics. 
}
\label{tab:corr}
\vspace{-2ex}
\end{table*}

\section{Experiments}

We investigate the extent to which automated metrics on an evaluation dataset correlate  with human ratings of system outputs. 
We use the human ratings collected by \citet{gupta2019investigating}, who collected utterance level human ratings using Amazon Mechanical Turk (AMT). They used a collection of 100 dialogue contexts that are randomly selected from the DailyDialog dataset. The generated response from various methods are rated in terms of appropriateness (from 1-5, with 5 denoting the best) by 5 different AMT workers. They collected and considered outputs from following methods: CVAE \cite{ZhaoZE17}, HRED \cite{Serban2016BuildingED}, Seq2Seq \cite{DBLP:journals/corr/VinyalsL15}, Dual-encoder \cite{DBLP:conf/sigdial/LowePSP15}, and Human-written responses. We report Spearman rank correlation \cite{spearman1961proof} and Kendall Tau rank correlation \cite{kendall1938new} of human ratings against ngram overlap metrics
such as BLEU \shortcite{papineni2002bleu}, METEOR \cite{banerjee2005meteor}, ROUGE-L \cite{lin2004rouge}, and embedding based metrics like cosine similarity of average word embedding (EmbeddingAvg) \cite{DBLP:journals/corr/WietingBGL15a} or Skip Thought Embedding \cite{DBLP:conf/nips/KirosZSZUTF15}, and precision (BERT-Prec) and recall (BERT-Rec) components of BertScore \cite{DBLP:conf/iclr/ZhangKWWA20}. 

We compare the correlations across following setups:
\noindent \textbf{\single{}} \cite{li-etal-2017-dailydialog}: Original DailyDialog dataset which had one reference per context; 
\textbf{\oursSingle{}}: Proposed method along with \single{} reference;
\textbf{\multiref{}} \cite{gupta2019investigating}: 4 human written references. 
\textbf{\oursMulti{}}: Reference responses from the proposed method along with \multiref{} references. 
Additionally, we report the results when using \paraphrase{} instead of \ours{}: \textbf{\paraphrase{}-\single{}} and \textbf{\paraphrase{}-\multiref{}}.
Paraphrasing is a popular approach for automated data augmentation.  
Paraphrasing via backtranslation (BT) \cite{sennrich-etal-2016-improving} is known to be an effective, domain-independent way to generate good quality paraphrases \cite{wieting2017paranmt}. 
We use the BT model from \cite{xie2020unsupervised} with its default hyperparameters to sample 5 paraphrases per human written reference 
\\


\vspace{-0.4\abovedisplayskip}
\noindent \textbf{Results:} We observe that most of the metrics show large improvements in correlations to human ratings for appropriateness when used along with \single{} or \multiref{} (Table \ref{tab:corr}). 
In fact, rank correlations across most of the metrics are better for \oursSingle{} compared to \multiref{}, even though former uses only single human written reference while latter uses upto 5 human written references\footnote{Rank correlations for \single{} and \multiref{} deviate from the values in \citet{gupta2019investigating}, who (in private communication with us)
, confirmed 
that the final dataset and code available on their repo does lead to the numbers we report. 
}. 
Additionally, we observe that \paraphrase{} produces little or no improvements in correlations with human ratings (Table \ref{tab:corr}). 
We posit that for a given response, alternate responses constitute a strictly richer subspace than that of response paraphrases, which tend to be lexico-syntactically variant but semantically invariant. \\

\begin{table}[t]
\centering
\small 
\begin{tabular}{@{}lcc@{}}
  Method &  \multicolumn{1}{c}{BLEU-4} & \multicolumn{1}{c}{BERT-Prec}  \\\toprule
\oursSingle{} & $0.30$ & $0.38$ \\ 
\midrule
\quad \oursSingle{} variants: \\ 
\commonsense{} only  & $0.24$ & $0.31$ \\ 
\textsc{Retrieval} only   & $0.29$ & $0.36$ \\
\textsc{Retrieval} only (5\% corpus) & $0.17$ & $0.28$ \\
\textsc{w/o context-adapt} & $0.26$  & $0.37$   \\ 
\bottomrule  
\end{tabular}
\vspace{-0.3\abovedisplayskip}
\caption{ 
Analyzing the impact of various components
}
\label{tab:ablations}
\vspace{-3ex}
\end{table}

\vspace{-0.4\abovedisplayskip}
\noindent \textbf{Analyzing the impact of various components:}
To understand the impact of various components, we report Spearman rank correlation scores for BLEU-4 and BERT-Prec metrics with some variants of \oursSingle{} (Table \ref{tab:ablations}). We note that considering only one knowledge source (\commonsense{}-only, \textsc{Retrieval}-only)  leads to good Spearman rank correlations of automated metrics to human ratings. Thus, the additive effect (\oursSingle{}) shows rather small incremental benefit. Moreover, \textsc{Retrieval} by itself does better than \commonsense{}, though at smaller corpus availability (e.g. 5\%), \commonsense{} performs better. 
Finally, not using context adaptation (\textsc{w/o context-adapt}) leads to significant performance drop. \\

\noindent \textbf{Qualitative Examples} To illustrate our approach, we present a couple of examples in Table \ref{tab:qualExamplesmain}. (A wider selection of examples can be found in Appendix Table 6.) \\

\noindent \textbf{Quality of Auto-generated References:}
We check the quality of \ours{} references by recruiting human annotators, showing them the reference along with the dialog context, and requesting them to tag each reference as appropriate, neutral, or not-appropriate, with respect to the dialog context. We randomly select $150$ responses each from \ours{} and \multiref{} for this purpose. 
We observe that in about $29\%$ of the references from \ours{} (fully automatically generated) were annotated as not appropriate, compared to $7\%$ for \multiref{}, demonstrating fair quality of augmented responses from \ours{} (Additional details and results in Appendix). We do note that the ones marked as not relevant/appropriate can often be tweaked easily by a human to transform them into valid responses -- demonstrating the possibility of exploring human-in-the-loop setups along with \ours{} to collect even better references.


\begin{table*}[ht!]
\centering
\small
\addtolength{\tabcolsep}{-4pt}
\begin{tabular}{ll}
Type  & Text \\ 
\toprule 
\textsc{Context} & {\color{black} A: How may I help you ?}   \\
\textsc{Single} & {\color{black} B: I'm having a problem.} \\
\hdashline
\textsc{Multi} & {\color{black} B: You can help me with this problem. || B: You can tell me how to get to customer service. }  \\  
 \textsc{Retrieval} & {\color{black} B: I have a problem. || B: There is a problem with my check. }  \\
\textsc{Commonsense} & {\color{black} B: I want to find information. || B: I want to ask question. || B: I want to make appointment.} \\ 
\toprule
\textsc{Context} & {\color{black} B: It can be solved by drawing a draft on us at 90 days sight.} {\color{black} A: What about a draft at 120 days sight ?}   \\
\textsc{Single} & {\color{black} B: All right. But we demand the draft be accepted by a bank acceptable to us.} \\
\hdashline
\textsc{Multi} & {\color{black} B: We'd like the matter resolved sooner. ||  B: We can do that, but there will be a higher interest rate.}  \\  
 \textsc{Retrieval} & {\color{black} B: Well, that's a lot of time to wait for the draft to be drawn.}  \\
\textsc{Commonsense} & {\color{black} B: I want to sign the contract. || B: I will look at the draft sheet.} \\  \toprule
\end{tabular}
\caption{Example showing the automated responses returned by different sub-components of \ours{}. Multiple responses from the same sub-component are separated by `||'. 
}
\label{tab:qualExamplesmain}
\vspace{-3ex}
\end{table*}

\section{Discussion}
\label{sec:discussion}

\noindent \textbf{Transferability to more languages:}
Transferability of our approach to more languages is one aspect that merits discussion. While commonsense resources aren't readily available in all languages, a workaround can be to use off-the-shelf MT to translate before querying into English versions of the commonsense resources, and then translate back retrieved information. 
Furthermore, we note that while commonsense knowledge was useful, removing the \textsc{Commonsense} method and relying on retrieval alone causes only relatively modest drop in performance (see Table \ref{tab:ablations}). Thus, for languages lacking commonsense resources, one may still attain good gains in reference based evaluation by retrieving and adapting from dialog corpus alone. \\

\noindent \textbf{Reference-less metrics:}
We note that while comparisons of using proposed approach against using reference-free metrics \cite{lowe2017towards,tao2017ruber} would be interesting, the focus of the current work is on improving reference-based evaluation via unsupervised reference augmentation.  
While reference-less metrics offer convenience to work with zero or a very small number of references, reference-based metrics can be advantageous on several fronts. Reference-based evaluation can be more interpretable under certain situations by identifying the reference which matches the most with a given system output. Reference-based evaluations allow for easy incorporation of additional references -- in contrast, many learned model-based metrics will require retraining if additional annotations become available. 

\section{Related Work}

Prior work explores many ways to improve over single-reference evaluation without collecting multiple ones.
\citet{fomicheva2020multi} obviate need for multiple references in MT by generating many ``alt-hypotheses" via test-time dropout from the same model. 
\citet{sai2020improving} and \citet{gupta2019investigating} collect additional manually annotated responses for dialog contexts. Compare to them, our method of automatically collecting additional references automatically is more scalable.

Automatic data augmentation in NLP has largely been used for increasing training data \cite{feng2020genaug,wei2019eda,feng2021survey}. In this work, we use retrieved dialog instances and commonsense knowledge base to augment reference set for a given dialog context. 
$\Delta$-Bleu \cite{galley2015deltableu} and uBLEU \cite{yuma2020ubleu} also use retrieval to produce pseudo-references for dialog response evaluation. Compared to $\Delta$-Bleu and uBLEU, our work is different since we utilize commonsense knowledge base and perform contextual adaptation.
Prior work in dialog response generation has explored the use of commonsense knowledge base \cite{emnlp2020persona} as well as retrieval \cite{song2016two,acl2021pabst} -- in contrast, our focus is on augmenting reference set for improving evaluation.

Automatic model-based metrics like ADEM \cite{lowe2017towards} and RUBER \cite{tao2017ruber}, which incorporate context while scoring for evaluation, at first glance seem to reduce the need for multiple references. However, these metrics have been found to suffer from several peculiar problems. For instance, ADEM can't discriminate between gold responses and certain classes of adversarial negatives e.g reversed gold responses and repeating the context as the response
\cite{sai2019re}. 
\citet{sato2020evaluating}  evaluate dialog systems through their ability at selecting valid responses from a semi-automatically curated candidate list.
\citet{mehri2020usr} introduce the unsupervised, reference-free USR metric, which leverages a suite of RoBERTa \cite{liu2019roberta} models, each finetuned to score one of five dialog aspects e.g \textit{Natural} and \textit{Uses Knowledge}. \citet{mehri2020unsupervised} further expand their USR metric to eighteen aspects from the initial five. 

\section{Conclusion}
In this work, we demonstrate how existing knowledge sources can be used to construct a diverse set of references in an automated and scalable manner. 
The resulting reference set demonstrates high correlation with human ratings of system outputs.

In future, we plan to incorporate other commonsense types into \ours, such as social \cite{sap2019social} and moral \cite{forbes2020social}. We also hope to explore human-in-the-loop setups which build on \ours{} to collect even better references.

\section*{Acknowledgements}
We thank anonymous ACL reviewers for insightful comments and feedback. We thank Prakhar Gupta \cite{gupta2019investigating} for useful discussions.

\section*{Ethics Statement}
Our preference ratings are collected over source content from an already existing, publicly available and widely used dataset i.e DailyDialog  \cite{li-etal-2017-dailydialog} 
%
We neither solicit, record or request any kind of personal or identity information while collecting our ratings.
%
Our work  primarily performs experiments on dialog in English \cite{bender2018data}.
Dialog models are known to suffer from biases learnable from dialog training data, such as gender bias \cite{dinan2019queens}. However, our work and contribution does not present or release any new models or model checkpoints, and is primarily focussed on making existing evaluation setups better through automated collection of larger reference sets.

\bibliography{references}
\bibliographystyle{acl_natbib}
\newpage

\clearpage
\appendix
\section{Additional Results}

\subsection{Additional Correlation Results}
\label{appendix:corr}

Table \ref{tab:corr_with_pvalues} shows Spearman rank correlation scores with p-values. 

\begin{table*}[t]
\centering
\small 
\begin{tabular}{@{}l||c|c|c||c|c|c@{}}
  \multicolumn{7}{c}{Spearman Rank Correlation (p-values)} \\ \toprule
  Setup  &  \multicolumn{3}{c||}{1 human written reference} &  \multicolumn{3}{c}{4 human written references}  \\ \midrule
    Dataset  & \multicolumn{1}{c|}{\single{}} & \paraphrase{} & \multicolumn{1}{c||}{\ours{}}  & \multicolumn{1}{c|}{\multiref{}} & \paraphrase{} & \multicolumn{1}{c}{\ours{}} \\ 
  & \multicolumn{1}{c|}{\tiny{\cite{li-etal-2017-dailydialog}}} & -\single{} & \multicolumn{1}{c||}{-\single{}(Ours)} & \multicolumn{1}{c|}{\tiny{\cite{gupta2019investigating}}} & -\multiref{} & \multicolumn{1}{c}{-\multiref{}(Ours)} \\ \midrule
BLEU-4 & 0.093 (0.04) & 0.135 (0.00) & 0.302 (0.00) & 0.281 (0.00) & 0.269 (0.00) & 0.357 (0.00) \\
BLEU-3 & 0.055 (0.22) & 0.105 (0.02) & 0.291 (0.00) & 0.243 (0.00) & 0.238 (0.00) & 0.345 (0.00) \\
BLEU-2 & 0.040 (0.37) & 0.082 (0.07) & 0.275 (0.00) & 0.203 (0.00) & 0.206 (0.00) & 0.327 (0.00) \\
BLEU-1 & 0.024 (0.59) & 0.062 (0.17) & 0.250 (0.00) & 0.191 (0.00) & 0.178 (0.00) & 0.295 (0.00) \\
ROUGE-L & 0.071 (0.11) & 0.088 (0.05) & 0.259 (0.00) & 0.197 (0.00) & 0.196 (0.00) & 0.317 (0.00) \\
METEOR & 0.106 (0.02) & 0.094 (0.04) & 0.243 (0.00) & 0.227 (0.00) & 0.217 (0.00) & 0.299 (0.00) \\
\midrule
EmbeddingAvg & 0.030 (0.50) & 0.015 (0.73) & 0.025 (0.58) & 0.099 (0.03) & 0.096 (0.03) & 0.079 (0.08) \\
SkipThought & -0.003 (0.95) & -0.033 (0.46) & 0.087 (0.05) & 0.065 (0.15) & 0.053 (0.24) & 0.129 (0.00) \\
BERT-Prec & 0.270 (0.00) & 0.279 (0.00) & 0.378 (0.00) & 0.319 (0.00) & 0.322 (0.00) & 0.407 (0.00) \\
BERT-Rec & 0.096 (0.03) & 0.094 (0.04) & 0.240 (0.00) & 0.232 (0.00) & 0.212 (0.00) & 0.304 (0.00) \\
\midrule
Max. value & 0.270 & 0.279 & 0.378 & 0.319 & 0.322 & 0.407 \\
\bottomrule
\end{tabular}
\caption{\small
Utterance level Spearman Rank Correlation \cite{spearman1961proof} with p-values. 
(1) \oursSingle{} augments the original single human written response (\single{}) in DailyDialog dataset \cite{li-etal-2017-dailydialog}  using proposed method. It leads to large improvements in correlations across most of the metrics, when compared to \single{}. 
(2) \oursMulti{} augments the \multiref{} dataset, again leading to improvements in correlations to human ratings, especially for BLEU and BERT-Prec metrics. 
Additionally, we note that almost all of the correlation values with \ours{}-\multiref{} are statistically significant with $p<0.05$.
}
\label{tab:corr_with_pvalues}
\end{table*}

\subsection{Quality Assessment based on RUBER}
As a second, automated way of ascertaining response quality, we use the unreferenced part of the RUBER metric \cite{tao2017ruber}, which uses a pretrained model to score quality of responses based on context alone. Here, we use the RUBER checkpoint\footnote{\url{tinyurl.com/ynqd54tt}} from \cite{sai2020improving}, which first pretrains on a large Reddit dataset, followed by finetuning on DailyDialog. \textsc{Single} and \textsc{Multi} have a quality of $\approx$ 0.72, while for 
\textsc{Retrieval} the values is $0.63$ . \textsc{CommonSense} is found to have the most superior quality at $0.82$, surpassing even \textsc{MULTI}. 


\subsection{Diversity of References}
\label{appendix:diversity}
We investigate the diversity of the references by computing self-BLEU scores \cite{zhu2018texygen} among references from \paraphrase{} vs \ours{}. For fair comparison, we randomly chose 4 references from corresponding method. We observe self-BLEU4 scores of $0.36$ for \paraphrase{} compared to only $0.13$\footnote{Note that lower self-BLEU denotes more diverse} for \ours{}.  

\section{Additional Details on Context Adaptation}
\label{appendix:context}

\subsection{Templates to convert Knowledge Base Outputs to Full Sentences}
\label{appendix:context-comet}

Table \ref{tab:templates} lists the set of templates and rules used to transform semi-structured \comet{} outputs to surface natural language forms.

\begin{table}[t]
\centering
\small
\begin{tabular}{@{}ll@{}}
Condition & Action \\
\toprule
Type is \textsc{oEffect} &  Prepend `I feel'  \\ 
Example: &   \\ 
\textsc{oEffect} (excited)  & => `I feel excited.' \\ 
\midrule
Type is \textsc{oWant} &  Prepend `I'  \\ 
Example: &   \\ 
\textsc{oWant} (to thank personx) &  => `I want to thank personx.' \\ 
\midrule
Type is \textsc{oReact} &  Prepend `I will'  \\ 
Example: &   \\ 
\textsc{oReact} (have a party) &  => `I will have a party.' \\ 
\midrule
Word \textsc{personx} &  Replace with `you'  \\ 
Example: &   \\ 
i thank \textsc{personx}. & => `I thank \emph{you}.' \\ 
\bottomrule
\end{tabular}
\caption{Templates and rules to transform semi-structured \comet{} outputs to surface NL forms.}
\label{tab:templates}
\vspace{-3ex}
\end{table}
\subsection{Unsupervised Decoding Procedure For Context Adaptation}
\label{appendix:context-delorean}
We use the author's own implementation\footnote{\url{tinyurl.com/2lqp9z6s}} of their DELOREAN decoding algorithm from \cite{DBLP:conf/emnlp/QinSWBHBBC20}. We use default hyperparameters from their implementation, which use the non-finetuned \textit{gpt2-medium} checkpoint as the LM atop which the unsupervised, gradient-based decoding procedure is run. Note that the model parameters are not updated in any way - the gradient computation and updates here are happening w.r.t the states, or more specifically, the state activation. 
More specifically, authors propose an iterature procedure wherein they alternatively perform forward and backward passes. In the forward pass, the current output $Y$ is updated as per the likelihood of the underlying decoder. In the backward pass, the output is updated to be as similar as possible to the sentence $Z$ from the knowledge source using back-propagation. However, since $Y$is discrete, we maintain a soft representation $\widetilde{Y}$ of the output $Y$ wherein $\widetilde{Y_i}$ represents the logits at the $i^{th}$ position as per the underlying decoder.
Next, we shall describe the backward and forward passes of the iterative procedure:

\textbf{1:} In backward pass, we update logits based on the gradients of a content-matching loss function $\triangledown_{\widetilde{Y}} L(\widetilde{Y}_{t-1},Z)$ giving backward logits $\widetilde{y}^{b}_{t}$ 

\textbf{2:} Next, we perform forward pass using the underlying decoder for steps $1$ to $N$. During forward pass at step $t$, we compute the logits $\widetilde{y}^{f}_{t}$ based on left context i.e. $X$ and $Y_{<t}$.  Next we perform weighted averaging of the forward and backward logits at step $t$ to arrive at the final logits to be used for the next time step in forward pass.

$\widetilde{Y_i}$ is initialized by performing a forward pass conditioned only on $X$ as per greedy decoding. We alternatively perform backward and forward passes till convergence. 
Final response is obtained via the resulting logit outputs $\widetilde{Y}$.



Specifically, we use their ``counterfactual'' setup, where an ending $e_{old}$ is adapted from its old context $c_{old}$ to an altered, new context $c_{new}$, generating a new, predicted ending $\hat{e}_{new}$. In our case, $c_{new}$ is the dialog context for the turn under evaluation $d^{past}_t$. In the \textsc{Retrieval} case, $c_{old}$ is the context of the retrieved candidate turn $x^{past}_{t'}$. For the \textsc{Commonsense} case, $c_{old}$ is also our current context, i.e the same as $c_{new}$ - we're simply attuning the already drawn inference better to the current context.

\section{Retrieval Similarity Function - Details}
\label{appendix:bm25_details}
Consider a dialog $d$ , broken up by turns as $\{C_{1}\ldots C_{t}, C_{t+1}{=} d^{resp}_{t}, C_{t+2} \ldots C_{T}\}$, where $t+1$ denotes the turn currently under evaluation. For the context-response $C^{1}_{t},\hat{r}_{t}$ pair to be evaluated, we retrieve pseudo-references based on a combination of of a) Past $d^{past}_{t}=C^{t-L_{b}}_{t}$ b) Gold response $d^{resp}_{t}$ c) Future $d^{future}_{t}=C^{t+2}_{t+2+L_{f}}$. $L_{b}$ and $L_{f}$ are past and future context windows.
Our retrieval similarity function is a sum of the log scores between each corresponding element of the turn under evaluation with the candidate turn.
\begin{center}
{}{\tiny
\begin{align*}
\vspace{-2ex}
    Sim(d_{t},x_{t'}) &=  \log S_{bm25}(d^{past}_{t},x^{past}_{t'})  + \log S_{bm25}(d^{resp}_{t},x^{resp}_{t'}) \\  &+ \log S_{bm25}(d^{future}_{t},x^{future}_{t'})
\vspace{-2ex}
\normalsize
\end{align*}}%
\end{center}
We set $L_{b}=L_{f}=2$ without specific tuning, as an intuitive tradeoff between enough specificity and enough possibility of relevant candidates.

BM25 \cite{robertson1995okapi} or ``Best Match 25'' is a tfidf like similarity. Its specific form is:
\begin{center}
\tiny
\begin{align*}
    S_{BM25}(q,d) &= \sum_{w_{i} \in q} \log(\frac{N}{{df}_{i}}) \frac{(k_{1}+1){tf}_{i}}{k_{1}((1-b)+b\frac{dl}{avdl})+{tf}_{i}}
\end{align*}
\normalsize
\end{center}
Here, ${tf}_{i}$ and $df_{i}$ are the term frequency in the current document and the document frequency (in the corpus). $N$ is corpus size, while $dl$ and $avdl$ are current and average document lengths. $b$ controls extent of document length normalization, while $k_{1}$ controls effect of term frequency. With $b=0$ and $k_{1}{}\rightarrow{}\infty$, this reduces to simple tfidf . Here, we use default gensim values, $b=0.7$, $k_{1}=0.5$
 \section{Qualitative Examples}
\label{appendix:qualExamples}
 In Tables \ref{tab:qualExamples}, we list some examples, each illustrating a turn of a test dialog with its immediate past, future, the four additional human references from \cite{gupta2019investigating} (shown under \textsc{Multi 2,3} and \textsc{Multi 4,5}), followed by automated response sets from different sub-components of \ours{}.
 
 \subsection{Before/After \textsc{ContextAdapt}}
 In Example \emph{4-4} of Table \ref{tab:qualExamples}, we can observe how \textit{``Yes , I'm young , and unmarried . It's no problem for me to travel frequently .''} gets context-adapted (shown as +\textsc{CA}, short for \textsc{ContextAdapt}) to \textit{``Yes , I'm able to understand English. It 's not that I don't understand English .''} which indeed does match the preceding dialog better. Similarly, in Example \emph{50-2} of Table \ref{tab:qualExamples}, we can see how \textit{``Well, that might be acceptable if you handle insurance fees"} is modulated stronger to the context which asks about duration, getting adapted to \textit{``Well, that's a lot of time to wait for the draft to be drawn."}. Note that we omit this row for the examples where it simply leaves the input unchanged, or produces outputs which are noticeably unfaithful or ill-formed.
 

\subsection{Complementarity of Components}
Sometimes, a component may suffer from example specific issues e.g In Example \emph{35-2}, the \textsc{Commonsense} approaches misinterpret what is a driving ticket in the context of the dialog as an event ticket, drawing inferences accordingly. However, even in such cases, the other component salvages the situation and ensures overall response set remains healthy - e.g, here, \textsc{Retrieval} produces pertinent responses like \textit{Could you tell me how you dealt with it this time?}, \textit{No I haven't. What about you ?} etc. In Example \emph{10-3}, we see the opposite situation, where the responses from \textsc{Retrieval} are somewhat less relevant, but \textsc{Commonsense} produces very pertinent responses such as \textit{``i decline the date"} and \textit{``I go on another date''}

\begin{table*}[ht!]
\centering
\scriptsize
\addtolength{\tabcolsep}{-4pt}
\begin{tabular}{|l|l|l| }
\hline 
No & Type  & Text \\ \hline 
\multirow{3}{*}{0-5} &   \textsc{Context} & {\color{red} A: I also have blow if you prefer to do a few lines.}  \\
& \textsc{Future} & {\color{teal} A: come on man ! i even got dope and acid ! try some !.}  \\
&  \textsc{Single} &    {\color{brown} B: No, I am ok, really.} \\
& \textsc{Multi 2,3} & {\color{violet} B: what is your price ? ; B: no thanks, I'm out of cash } \\  
& \textsc{Multi 4,5} & {\color{violet} B: yeah , what have you go got ?; B: maybe tomorrow , i 'm running late .} \\ 
& \textsc{Retrieval} & {\color{blue} B: Well , I guess Zina figured you couldn't blow it if you were in the hotel room .} \\
& \textsc{Retrieval w CA} & {\color{blue} B: I have a few lines of blow , but it is not necessary in the way of the rest of the script .} \\
& \textsc{Commonsense} & {\color{orange} B: i want to go to bathroom; B: i want to play game; B: i want to have rest; B: i want to sleep} \\
\hline
\multirow{3}{*}{1-1} &   \textsc{Context} & {\color{red} B: The taxi drivers are on strike again A: What for?}  \\
& \textsc{Future} & {\color{teal} A: it is really a hot potato .}  \\
&  \textsc{Single} & {\color{brown} B: They want the government to reduce the price of gasoline .} \\
& \textsc{Multi 2,3} & {\color{violet} B: they want more pay for their time .; B: they are not happy with their permit related expenses .}  \\  
& \textsc{Multi 4,5} & {\color{violet} B: they want healthcare and benefits .; B: i don't know . i just know there are no cabs in the city} \\ 
& \textsc{Retrieval} & {\color{blue} B: Well . Don't worry about it , I will be happy to share the taxi with you .} \\
&  & {\color{blue} B: I think the price is a little high , can't you reduce it?} \\
& \textsc{Retrieval w CA} & {\color{blue} B: Well , they are not going to be able to get their fares back .} \\
& & {\color{blue} B: I think the reason is a little more complicated than that .} \\
\hline
\multirow{3}{*}{4-4} &   \textsc{Context} & {\color{red} A: Do you think you can make yourself easily understood in English ?}   \\
&  \textsc{Single} & {\color{brown} B: yes , in most circumstances .} \\
& \textsc{Multi 2,3} & {\color{violet} B: my english is fine .   my keyboard sometimes just acts up . ;}  {\color{violet} B: yes, and   i'm constantly trying to improve my communication.} \\  
& \textsc{Multi 4,5} & {\color{violet} B: i do n't know .   i may be to nervous and then i 'll mess up . ;}  {\color{violet} B: yes , i believe so .   i 've been speaking english since birth .} \\ 
& \textsc{Retrieval} & {\color{blue} B: Yes , I'm young , and unmarried . It's no problem for me to travel frequently .} \\
& \textsc{Retrieval w CA} & {\color{blue} B: Yes , I'm able to understand English . It 's not that I don't understand English .} \\
\hline
\multirow{3}{*}{10-3} &   \textsc{Context} & {\color{red}B: Hold on , please . Let me check it for you . Yes , you're right . You will keep it for 3 days .} \\&& {\color{red} A: Well , now I want to change the date from 24th to 28th . }  \\
&  \textsc{Single} & {\color{brown} B: ok , that shall be arranged .} \\
& \textsc{Multi 2,3} & {\color{violet} B: we can easily do that for you . ;  B: i 'm sorry but someone's reserved the room for 28th .   i can offer you a different room ?} \\  
& \textsc{Multi 4,5} & {\color{violet} B: i can extend your stay time but it may cost more since it is going into the holiday . ; } \\ & & {\color{violet} B: i can give you a different double room for the 28th at a discounted price .} \\ 
& \textsc{Retrieval} & {\color{blue} B: All right . May I have your name ? ; B: Apparently there is only \$ 57 left in your account . ;} \\&& {\color{blue} B: Here you are . What is the rate ?; B: I prefer not to move first .} \\
& \textsc{Commonsense} & {\color{orange} B: i decline the date ; B: i go on another date; B: i go on a date;} \\&&  {\color{orange} B: i get nervous; B: i get rejected; B: i get dumped; B: i feel sad} \\ 
\hline
\multirow{3}{*}{23-10} &   \textsc{Context} & {\color{red} B: I mainly eat baked chicken , because there's not a lot of fat . } {\color{red}A: that does sound pretty good .}   \\
&  \textsc{Single} & {\color{brown} B: i know it does , and that 's because it is .} \\
& \textsc{Multi 2,3} & {\color{violet} B: yes , it is very good and pretty healthy .;  B: it is good , you should come over and i will make you dinner .}  \\  
& \textsc{Multi 4,5} & {\color{violet} B: thanks , it 's been a struggle , but i am starting to see results . ;} \\&& {\color{violet} B: i appreciate your kind words , i am getting used to eating more healthy .} \\ 
& \textsc{Retrieval} & {\color{blue} B: You should try it . You won't be disappointed .;  B: Do you have any plans for tomorrow ? ; B: That's it ? ; } \\&& {\color{blue} B: Just about . ; B: I like fish . Please bring me an order of that .} \\
& \textsc{Commonsense} & {\color{orange} B: i want to cook dinner; B: i want to go to movie; B: i want to write letter ;} \\ && {\color{orange} B: i will open your mouth; B: i will think about what you want; B: i will drink some water; } \\ 
\hline
\multirow{3}{*}{35-2} &   \textsc{Context} & {\color{red} B: I've never gotten one . Have you ?}  {\color{red} A: I got a ticket yesterday .}   \\
& \textsc{Future} & {\color{teal} A: i parked in the red zone .} \\
&  \textsc{Single} & {\color{brown} B: really ? what did you get one for ?} \\
& \textsc{Multi 2,3} & {\color{violet} B: what was it for ? ;  B: how much is it ?}  \\  
& \textsc{Multi 4,5} & {\color{violet} B: do you resent you got a ticket ? ;  B: are you going to appeal the ticket ?} \\ 
& \textsc{Retrieval} & {\color{blue} B: I've gotten a few .; B: No , I haven't . What about you ? ;} \\&& {\color{blue} B: Could you tell me how you dealt with it this time ? ;} \\&& {\color{blue} B: I will explain it in detail in the next class , could you preview it ?}  \\
& \textsc{Commonsense} & {\color{orange} B: i give ticket to you; B: i give ticket to the cashier; B: i give ticket to the clerk.; } \\&& {\color{orange} B: i feel happy .; B: I feel excited .; B: i feel annoyed . ; B: I feel disappointed .; } \\&&  {\color{orange} B: i see what else they can do .; B: i see what else they can do to get the ticket; i go to the event } \\ 
\hline
\end{tabular}
\caption{Example context-response pairs from the test split of DailyDialog, showing the automated responses returned by different sub-components of \ours{}. \textsc{ContextAdapt} is shortened to \textsc{CA} for brevity.}
\label{tab:qualExamples}
\vspace{-3ex}
\end{table*}

\section{Human Evaluation Details}

\subsection{Quality of References}
The quality of references were judged by two graduate students from a university where the medium of instruction is English. The annotators were requested to ignore minor grammar issues, and focus more on the content of the response. 

\section{Computing Details}
The GPUs used for \textsc{Commonsense} and \textsc{ContextAdapt} experiments were a Geforce Rtx 2080 and TitanX Pascal respectively.

\clearpage

\end{document}